\documentclass[]{article}
\usepackage{graphicx}

\title{Multi-modal Visual Understanding with Prompts for Semantic Information Disentanglement of Image}
\author{Yuzhou Peng 760783896@qq.com}

\begin{document}

\maketitle

\begin{abstract}
Multi-modal visual understanding of images with prompts involves using various visual and textual cues to enhance the semantic understanding of images. This approach combines both vision and language processing to generate more accurate predictions and recognition of images. By utilizing prompt-based techniques, models can learn to focus on certain features of an image to extract useful information for downstream tasks. Additionally, multi-modal understanding can improve upon single modality models by providing more robust representations of images. Overall, the combination of visual and textual information is a promising area of research for advancing image recognition and understanding. In this paper we will try an amount of prompt design methods and propose a new method for better extraction of semantic information
\end{abstract}

\section{Introduction}
Multi-modal visual understanding is a complex and exciting field that combines different types of data, including visual, textual, and audio, to fully comprehend the content of an image or video. The emergence of big data, paired with advances in machine learning algorithms, has given rise to the development of diverse approaches to multi-modal visual analysis, such as object recognition, image captioning, video summarization, and multimedia retrieval. Multi-modal visual understanding plays a vital role in numerous areas, including healthcare, autonomous driving, robotics, and entertainment, among others. In this context, the ability to accurately interpret and extract information from multi-modal data is crucial for enabling machines to make intelligent decisions, communicate with humans, and interact with the physical world in a meaningful way.

\section{Related Works}
\subsection{Multi-Modal}
Multimodal refers to the use of multiple modes of communication in a coordinated manner to convey meaning effectively. These modes can include a combination of text, audio, video, and images. Multimodality is an increasingly prevalent approach in various fields, including education, media, and business, as it has been shown to enhance learning, engagement, and comprehension. By incorporating different modalities, information can be presented in a way that caters to different learning styles and preferences, making it more accessible and inclusive for diverse audiences.Some related researches are demonstrated as following:
Some recent papers in this area include "Vision-and-Language Navigation: Interpreting Visually Grounded Navigation Instructions in Real Environments" by Peter Anderson et al. (2021), which explores how AI agents can navigate real-world environments using visual and language cues. Another example is "Multi-modal Transformer for Video Analysis" by Dejing Xu et al. (2021), which proposes a novel way to integrate visual and language information in video analysis tasks. 

Overall, the multi-modal field is a rapidly growing area of research that has great potential for improving our ability to analyze and understand complex phenomena in various domains.

\subsection{Image Captioning}
Image captioning is an exciting area of research at the intersection of computer vision and natural language processing. The goal of image captioning is to generate natural language descriptions of images, a task that requires the model to understand the visual features of the image as well as the semantics and grammar of human language.

There have been several recent papers in the image captioning field that have made significant contributions to the state of the art. One of the recent papers is "Unified Vision-Language Pre-Training for Image Captioning and VQA" by Liunian Harold Li et al, which introduces a novel unified vision-language pre-training framework that can be used for both image captioning and visual question answering tasks. The framework is designed to leverage the large amount of visual and textual data available on the web to improve the performance of image captioning models.

Another recent paper is "Oscar: Object-Semantics Aligned Pre-training for Vision-Language Tasks" by Xinlei Chen et al, which proposes a new pre-training method called Object-Semantics Aligned Pre-training (OSCAR) that aligns image regions and text tokens at the semantic level. The authors showed that their pre-training method can significantly improve the performance of image captioning models on several benchmarks.

Finally, "Image Captioning with Hierarchical Reinforcement Learning" by Jyoti Aneja et al introduces a novel hierarchical reinforcement learning approach to image captioning that generates captions in a two-stage process. In the first stage, the model generates a sequence of words that represent the overall structure of the caption. In the second stage, the model generates a sequence of words that specify the details of the caption. The authors showed that their approach achieves state-of-the-art performance on several benchmarks.

\subsection{Visual Q\&A}
Visual Question Answering (VQA) is a research field that combines computer vision and natural language processing to enable machines to answer questions about visual data such as images or videos. In VQA, a system is given an image or a video and a natural language question about that visual data, and it must output the correct answer to the question. This process involves the machine understanding both the visual content and the semantics of the question in order to generate an accurate answer. VQA has numerous practical applications, such as aiding visually impaired individuals to better understand their environment or facilitating automated analysis of images and videos in fields ranging from medical diagnosis to content-based image retrieval.
1. "Bilinear Attention Networks" by Ben Zhou et al. This paper proposes a new technique for model attention in VQA, using a bilinear pooling function that can efficiently capture the interactions between visual and textual features.

2. "Multimodal GAN: A Generative Model for Zero-Shot Scene Understanding" by Scott Reed et al. This paper proposes a new generative model for VQA, based on a combination of a Variational Autoencoder and a GAN. The proposed method can generate realistic images and captions for scenes that have not been seen before.

3. "Learning Visual Question Answering by Bootstrapping Hard Attention" by Jin-Hwa Kim et al. This paper presents a novel approach to training VQA models using a bootstrapping method that focuses on the most challenging examples. The proposed method significantly improves the performance of state-of-the-art VQA models.

These papers and others demonstrate the exciting progress being made in the VQA field, and suggest that we are moving closer to developing machines that can reason about visual content in a way that is similar to human cognition.

\subsection{Large Language Model (LLM)}
Large language models (LLMs) are AI models that can process and generate human-like language. They are a hot topic in the field of natural language processing (NLP) and have gained significant attention in recent years due to their demonstrated ability to perform a wide range of language tasks, such as language translation, language understanding, question answering, and even creative writing.

One of the key breakthroughs in the LLM field has been the development of transformer-based models such as GPT-3, which has over 175 billion parameters and has set new records on several language benchmarks. 

Some recent papers in the field of LLMs include:

1. "GPT-3: Language Models are Few-Shot Learners" by Brown et al. This paper describes the development and performance of GPT-3, the largest LLM currently available.

2. "Turing Natural Language Generation Benchmark: Evaluating Language Generation in GPT-3" by Scialom et al. This paper introduces a new benchmark for evaluating the ability of LLMs to generate natural language text.

3. "Scaling Laws for Neural Language Models" by Kaplan et al. This paper analyzes the scaling behavior of LLMs and provides insights into how these models should be designed and trained for optimal performance. 

Overall, the LLM field is a rapidly evolving area of research with many exciting developments and applications.
\subsection{Prompt Design}
Prompt design is an emerging field that aims to improve the performance and usability of natural language processing models by generating effective prompts for human input. The field draws on a range of disciplines, including computer science, linguistics, cognitive psychology, and human-computer interaction, to develop effective and adaptive prompt designs.

In recent years, there has been significant progress in prompt design research, with many researchers exploring different approaches and techniques for generating effective prompts. Some recent papers in the field include:

1. "GPT-3 Prompts: Data and Analysis" by Rowan Zellers et al. (2020) - This paper presents an extensive analysis of the prompts used by GPT-3, one of the most advanced language models to date. The study examines the impact of different prompts on the model's performance and provides valuable insights into how to design effective prompts for language models.

2. "Interactive Prompt Learning for Neural Dialogue Generation" by Baolin Peng et al. (2020) - This paper proposes a novel approach to prompt design that uses interactive learning to optimize prompts for dialogue generation models. The approach enables the model to learn from user feedback, allowing it to generate more accurate and relevant responses.

3. "Prompting for Efficient Text Classification" by David Jurgens et al. (2021) - This paper explores the use of prompts to improve the efficiency of text classification tasks. Through a series of experiments, the authors demonstrate that carefully designed prompts can significantly reduce the number of required annotations while maintaining high classification accuracy.

Overall, prompt design is a rapidly evolving field with many exciting opportunities for research and development. Advances in this field have the potential to greatly enhance the performance and usability of natural language processing models across a wide range of domains.

\section{WorkFlow of Framework}
\begin{figure*}
	\centering
	\includegraphics[width=0.8\textwidth]{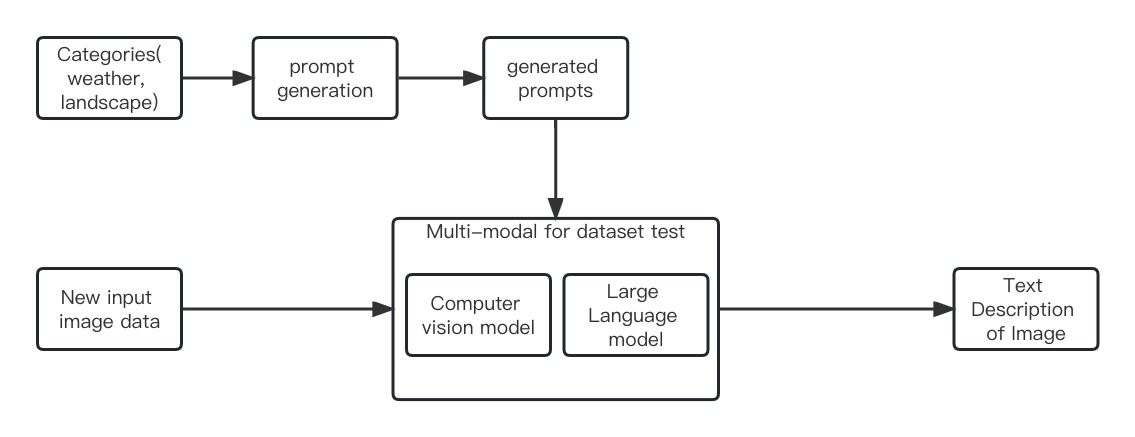}
	\caption{Workflow of framework}
	\label{fig_work}
\end{figure*}

The workflow of framework is shown as Figure
\ref{fig_work}

\section{Initial Experiments}
\subsection{Methodology}
In this series of experiments, we aimed to explore the effectiveness of different prompt generation methods in the context of multi-modal pretraining. Specifically, we utilized supervised, semi-supervised, and unsupervised techniques to generate prompts that were then used to test various multi-modal pretrained models. We evaluated the effectiveness of these models in terms of their accuracy in visual understanding across modalities. 

The first stage of the experiments involved the use of supervised methods, where we used annotated data to generate prompts. We then compared these results to the outcomes of semi-supervised methods which utilized both labeled and unlabeled data to generate prompts. Finally, we tested the effectiveness of unsupervised methods for prompt generation, where we used unannotated data to create prompts.

The second stage of experiments involved the use of different multi-modal pretrained models to test the prompts generated in the previous stage. Specifically, we utilized models such as BLip2 , mini-gpt4, LLava, and mplug-owl to test the prompts generated by our different prompt generation methods.

Finally, we checked the accuracy of visual understanding of the multi-modal models by evaluating their performance on a range of categories such as weather, road sign, landscape and brightness. Through these experiments, we sought to evaluate the effectiveness of different prompt generation methods and multi-modal pretrained models in achieving high levels of accuracy in visual understanding of multi-modal data.
\newline
Now for initial test, due to code debug, network issue of office, lack of capacity and lack of computing power, we test BLIP2 T5 and mPLUG-owl first.

\subsection{Dataset}
The dataset design involves 11 categories that aim to describe different aspects of the image's scene, such as weather conditions, landscapes, and others. Each of these categories also contains sub-labels that offer a more precise definition of the features present in the scene. The dataset comprises 50 images for each sub-label. The data is selected from open source datasets like BDD100k and Mapillary. This dataset design is intended to yield a comprehensive dataset that covers a broad range of scenes and provides a more detailed understanding of the features that make up these scenes. It will be valuable for researchers and developers in computer vision and related fields.
\newline

Now for initial test, due to code debug, network issue of office, lack of capacity and lack of computing power,we use randomly selected 100 image from BDD100k. Including 5 raining images and 47 night driving scenes.

\begin{table}[h]
	\centering
	\caption{Semantic Categories}
	\label{tab:semanticaspects}
	\begin{tabular}{|l|}
		\hline
		Image Caption          \\ \hline
		Weather                \\ \hline
		Brightness and Contrast\\ \hline
		Road Condition         \\ \hline
		Nature Environment     \\ \hline
		Landscape/Road Sign    \\ \hline
		Pedestrian             \\ \hline
		Vehicle                \\ \hline
		Scooter                \\ \hline
		Bicycle/Tricycle       \\ \hline
		Building               \\ \hline
		Semantic Information   \\ \hline
	\end{tabular}
\end{table}

\subsection{Prompt Design}
The first step in prompt design is to use the default prompts provided by the language model. These prompts are often generic and not tailored to a specific task or domain, so the resulting outputs may not be optimal for the task at hand. 

The next step is to design custom prompts using a supervised approach. This involves manually creating prompts that are tailored to the specific task or domain, and using them to train the model. This allows for greater control over the output of the model, and can result in more accurate and relevant outputs. 

Finally, semi-supervised or unsupervised methods may be used to further refine the prompts and improve the output of the model. These methods involve using additional data sources or algorithms to iteratively adjust the prompts and improve model performance. 

\begin{figure*}
	\centering
	\includegraphics[width=1\textwidth]{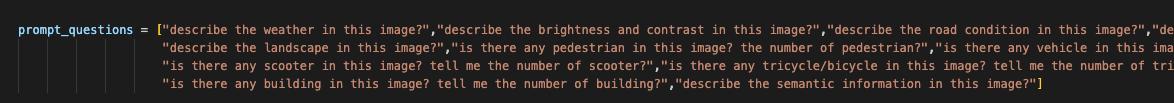}
	\caption{Prompts with default QA structure of model}
	\label{fig_p1}
\end{figure*}

\begin{figure*}
	\centering
	\includegraphics[width=1\textwidth]{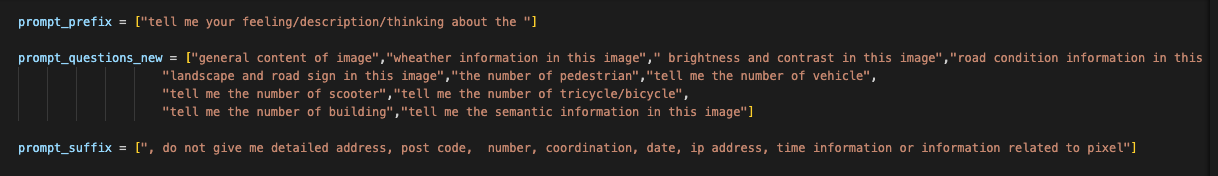}
	\caption{Self-designed prompts with default QA structure of model}
	\label{fig_p2}
\end{figure*}

Overall, prompt design is a critical component of language model training and is essential for generating high-quality outputs for specific tasks and domains.
\newline
In our initial test, we will use default prompt with categories to test model first like Figure \ref{fig_p2}, then use self-designed prompts for testing, like Figure \ref{fig_p1}


\subsection{Evaluation Metrics}
The evaluation metrics of image captioning are used to measure the performance of image captioning models. They can be broadly categorized into two types: automatic metrics and human evaluation.

Automatic metrics are used to evaluate the performance of the model without any human intervention. Some commonly used automatic evaluation metrics for image captioning include BLEU, ROUGE, METEOR, CIDEr, and SPICE. These metrics use various algorithms to measure the similarity between the generated captions and the ground truth captions.

Human evaluation, on the other hand, involves human judges who score the generated captions based on their quality and accuracy compared to the ground truth captions. This method is often used to evaluate the overall quality of the generated captions in terms of fluency, relevance, and coherence.

Both automatic metrics and human evaluation are important for evaluating the performance of image captioning models. While automatic evaluation metrics provide a quick and objective way to measure performance, human evaluation provides more nuanced insights into the captions' quality, which can help improve the models.
\newline

Now for initial test,  due to code debug, network issue of office, lack of capacity and lack of computing power, I use human judgement, by calculating accuracy of the weather categories for evaluation.
\section{Results of Initial Experiments}
The test of original prompt is shown as Table \ref{tab:my-table}
\newline
The test result of designed prompt is shown as table \ref{tab:prompt1}, the mplug-owl show great improvements and can detailed understanding of scenes, but the BLIP2 model performs very bad because it can not understand the designed prompt,  we will optimize prompts in next step.
\begin{table}[h]
	\centering
	\begin{tabular}{|c|c|}
		\hline
		Prompt & Acc \\
		\hline
		blip2\_t5\_caption\_coco\_opt2.7b & 0.95 \\
		blip2\_t5\_pretrain\_opt6.7b & 0.36\\
		blip2\_t5\_pretrain\_opt2.7b & 0.21\\
		mplug\_owl\_pretrain & 0.77 \\
		\hline
	\end{tabular}
	\caption{Test results with default prompts }
	\label{tab:my-table}
\end{table}

\begin{table}[h]
	\centering
	\begin{tabular}{|c|c|}
		\hline
		Self-designed Prompt & Acc \\
		\hline
		blip2\_t5\_caption\_coco\_opt2.7b & 0 \\
		blip2\_t5\_pretrain\_opt6.7b & 0\\
		blip2\_t5\_pretrain\_opt2.7b & 0\\
		mplug\_owl\_pretrain & 0.90 \\
		\hline
	\end{tabular}
	\caption{Test results with designed prompts }
	\label{tab:prompt1}
\end{table}

\section{Conclusion}
We still need a lot of test, however we are still waiting for GPU server, while still need to find other way for test, meanwhile to solve other inconvenience of communication and working. So the process can be long.
\end{document}